\documentclass[11pt]{article}
\usepackage{times}
\usepackage{graphicx}
\usepackage[hidelinks]{hyperref}
\usepackage{latexsym}
\usepackage[margin=4cm,columnsep=.8cm]{geometry}

\usepackage[ruled,noline,linesnumbered]{algorithm2e}
\usepackage{amsmath}
\usepackage{amssymb}
\usepackage{comment}
\usepackage{float}
\usepackage{graphicx}
\usepackage{soul}
\usepackage{url}
\usepackage{xcolor}
\usepackage{subcaption}
\let\oldnl\nl
\DeclareMathOperator*{\argmax}{argmxax}
\newcommand{\nonl}{\renewcommand{\nl}{\let\nl\oldnl}}

\newcommand{\myvec}[1]{\mathbf{#1}}
\newcommand{\V}[1]{\text{Var}\left[#1\right]}

\newcommand{\Cov}[1]{\text{Cov}\left[#1\right]}

\begin{document}

\title{Fleet Control using Coregionalized Gaussian Process Policy Iteration}

\author{Timothy Verstraeten \and Pieter JK Libin \and
 Ann Now\'e}
\date{\small Artificial Intelligence Lab Brussels,\\Vrije Universiteit Brussel,
Belgium,\\\{tiverstr, plibin, anowe\}@vub.be}

\maketitle

\begin{abstract}
In many settings, as for example wind farms, multiple machines are instantiated to perform the same task, which is called a fleet. The recent advances with respect to the Internet of Things allow control devices and/or machines to connect through cloud-based architectures in order to share information about their status and environment. Such an infrastructure allows seamless data sharing between fleet members, which could greatly improve the sample-efficiency of reinforcement learning techniques. However in practice, these machines, while almost identical in design, have small discrepancies due to production errors or degradation, preventing control algorithms to simply aggregate and employ all fleet data. We propose a novel reinforcement learning method that learns to transfer knowledge between similar fleet members and creates member-specific dynamics models for control. Our algorithm uses Gaussian processes to establish cross-member covariances. This is significantly different from standard transfer learning methods, as the focus is not on sharing information over tasks, but rather over system specifications. We demonstrate our approach on two benchmarks and a realistic wind farm setting. Our method significantly outperforms two baseline approaches, namely individual learning and joint learning where all samples are aggregated, in terms of the median and variance of the results.
\end{abstract}

\section{INTRODUCTION}

Reinforcement Learning (RL) is a framework for optimizing control policies through trial-and-error \cite{intro_rl}. A learning agent operates within an environment and optimizes its control actions based on gathered experiences. As state-of-the-art techniques typically require a large amount of experiences, a key challenge in RL is to increase sample-efficiency, such that RL techniques become viable in real-life applications \cite{sample_efficiency_2018}.

In this work, our objective is to improve sample efficiency in the context of a \emph{fleet}, i.e., a set of machines instantiated to perform the same task and managed as a single system.\footnote{Although the word ``fleet'' usually refers to a group of ships or cars, it recently has been adopted by other fields in the context of any type of machinery.} Fleets are prominent in many industrial applications, such as wind farms \cite{wind_farms_fleet_2016,verstraeten2019fleetwide} and autonomous vehicles \cite{fleet_vehicles_2014}, because fleets are cheaper to maintain and operate. To this end, we present a method that aggregates the experiences of the distinct fleet members, in contrast to the experiences of a single machine. The time is right for such a method, as the recent advances in the \emph{Internet of Things} allow fleet members to share data from modern wireless sensors using a cloud-based architecture, rapidly providing a complete overview of the problem \cite{fleet_wide_condition_monitoring_2018}.

As fleet members carry out the same task, they typically share the same design. In reality, fleet members differ slightly in terms of dynamics, for example due to production errors or degradation \cite{degradation_wind_farm_2014}. Thus, naively aggregating data over all members is insufficient and can be detrimental for the learning process. Therefore, information should only be shared between fleet members that are sufficiently similar.

We propose a new RL method for fleet control where knowledge transfer over dynamics models of similar devices is possible without compromising the specificity of an individual's model. More specifically, we create a Bayesian RL method that uses a Gaussian process (GP) to model the dynamics for a single member and aims to estimate correlations with other members using a novel sparse coregionalization method.

GPs are Bayesian models known to successfully capture complex non-linear surfaces using only a limited amount of data. They have previously been used in the context of RL \cite{gprl,gptd,pilco} and are popular when high sample-efficiency is required. Coregionalization was originally introduced in geostatistics to generate valid covariance matrices for modeling multivariate data sets \cite{geo_coreg}. It has later been used in the context of multi-task learning to describe correlations between a set of tasks \cite{multitask_kernel}.

In this work, we develop a fleet-wide policy iteration method based on coregionalized GPs. We start by positioning our research within the literature (see Section~\ref{sec:related_work}) and provide background on GPs (Section~\ref{sec:gp}) and RL (Section~\ref{sec:rl}). Then, we describe the Bayesian fleet transition model and how a fleet member can access its specific predictive statistics (Section~\ref{sec:methods}). Next, we analyze the sample efficiency and performance of our method on fleet-variants of well-known benchmark settings, namely, the continuous mountain car and the underactuated cart-pole (Section~\ref{sec:experiments}). Additionally, we demonstrate the practical benefits of our method on a state-of-the-art wind farm simulator (Section~\ref{sec:experiments_wind}). Finally, we formulate conclusions and identify future work (Section~\ref{sec:conclusion}).

\section{RELATED WORK}
\label{sec:related_work}

The type of learning we consider in our work is related to multi-task (or inductive transfer) reinforcement learning \cite{pan2009survey}, where a set of control tasks is jointly learned, leveraging potential similarities between them. In contrast to our setting, most work on multi-task RL considers different task parameters, while the system specifications remain the same \cite{ijspeert2003learning,bayesian_mtrl_value,taylor2007cross,bayesian_mtrl_mdp,kober2011reinforcement,konidaris2012transfer,deisenroth2014multi}. Specifically, \cite{bayesian_mtrl_value} and \cite{bayesian_mtrl_mdp} construct a Bayesian (hierarchical) structure of tasks, where the task parameters are assumed to be drawn from a set of priors shared among similar tasks.
Recent work has focussed on MDPs for which the system specifications are different, but the reward function remains the same \cite{hidmdp_2016,hidden_param_mdp_2017,meta_rl_2018}. Typically, a latent embedding of the system specifications is learned in order to share information among various machines.

Our work is different as it considers fleet settings in which we assume that the system specifications are nearly identical up to degraded parts or small design discrepancies. This means that a more targeted approach can be taken. Rather than having a single latent embedding from which all members originate, a more directed peer-to-peer transfer method can be taken through correlations. Such direct transfer is sample-efficient, as estimating the fleet-wide correlations for a given target is limited to learning a set of parameters linear in the size of the fleet.

Fleet settings are inherently multi-agent systems. While multi-agent reinforcement learning deals with control and coordination and control in multi-agent systems \cite{marl_survey_2006}, it focuses on coordination problems, rather than information exchange between agents.


\section{GAUSSIAN PROCESS}
\label{sec:gp}

Gaussian processes (GPs) \cite{intro_gps} are an extension of multivariate normal distributions. Similar to the latter, a GP describes a set of normally distributed random variables that are potentially correlated, i.e., knowledge about one variable gives information about another. However, the difference with multivariate normal distributions is that a GP is defined over arbitrary sets of annotated random variables. In a regression context, these random variables are the outputs of an unknown function, and their annotations are the inputs to that function.

Formally, assuming a zero-mean GP prior, i.e.,
\begin{equation}
f(\myvec{x}) \sim \mathcal{GP}\left(0, k(\myvec{x}, \myvec{x'})\right),
\label{eq:gp}
\end{equation}
and any arbitrary set of inputs $X$, we can model the associated latent function values $\mathbf{f}$ as
\begin{equation}
\mathbf{f}\ |\ X \sim \mathcal{N}(\myvec{0}, K)
\label{eq:normal}
\end{equation}
where $K_{ij} = k(\myvec{x}_i, \myvec{x}_j)$ is the covariance between variables $f_i$ and $f_j$. When regressing over a training set $(X_{\text{tr}}, \myvec{y}_{\text{tr}})$, we can compute the posterior statistics of $(\myvec{f}\ |\ X, X_{\text{tr}}, \myvec{y}_{\text{tr}})$ to obtain the predictive outputs $\myvec{f}$ for inputs $X$.
For the zero-mean GP described in Equation~\ref{eq:normal}, we have:
\begin{equation}
\begin{split}
\mathbb{E}\left[\myvec{f}\ |\ X, X_{\text{tr}}, \myvec{y}_{\text{tr}}\right] &= K_{X,X_\text{tr}} C^{-1}_{X_\text{tr}, X_\text{tr}} \myvec{y}_\text{tr}\\
\mathbb{V}\left[\myvec{f}\ |\ X, X_{\text{tr}}, \myvec{y}_{\text{tr}}\right] &= K_{X,X} - K_{X,X_\text{tr}} C^{-1}_{X_\text{tr}, X_\text{tr}} K_{X_\text{tr}, X}\\
C_{X_\text{tr}, X_\text{tr}} &= K_{X_\text{tr}, X_\text{tr}} + \sigma^2 I,
\label{eq:post_normal}
\end{split}
\end{equation}
where $K_{X,X_\text{tr}}$ is a matrix containing the pair-wise covariances between sets $X$ and $X_\text{tr}$ according to the covariance kernel and $\sigma^2$ is observational noise.

The choice of covariance kernel $k(\cdot, \cdot)$ is important, as it defines the various characteristics about how the model should generalize from the training set. We use the squared exponential (SE) kernel, defined as:
\begin{equation}
k^{SE}_{\theta^{SE}}(\mathbf{x}, \mathbf{x'}) = \exp\left(-\sum^D_{d=1}\frac{(x_d - x'_d)^2}{2l_d^2}\right)
\label{eq:se_kernel}
\end{equation}
where $\theta^{SE}$ contains the hyperparameters $l_d$, which denotes the length scale along dimension $d$ and characterizes the smoothness of the unknown function. 
This kernel has several properties, including continuity, differentiability and stationarity, rendering it a popular choice for general modeling purposes. 

\section{REINFORCEMENT LEARNING}
\label{sec:rl}

Consider the Markov decision process (MDP) $\mathcal{M} = (S, A, \tau, \gamma, R)$ \cite{mdp_definition}. $S$ and $A$ are state and action spaces, respectively. The transition function $\tau(\myvec{s}, \myvec{a})$ returns the state $\myvec{s'}$ when executing action $\myvec{a}$ in state $\myvec{s}$. The reward function $R : S, A, S \to \mathbb{R}$ returns the immediate immediate reward. The discount factor $\gamma \in [0, 1)$ determines the importance of future rewards. Additionally, consider a policy $\pi: S \to A$, which defines how an agent behaves given a particular state.

We specify the reward function as a square-exponential\footnote{
The reward function can be learned using a GP without jeopardizing the analytical benefits of our method. However, as our work focuses on learning over multiple transition models, we assume a known reward function centered around a prespecified goal state.} centered around a goal state with width $\sigma_R$, i.e.,
\begin{equation}
R(\myvec{s}, \myvec{a}, \myvec{s'}) = \frac{1}{\sqrt{2\pi \sigma_R}}\exp\left(-\frac{||\myvec{s'} - \myvec{s}_\text{goal}||\strut^2_2}{2\sigma^2_R}\right).
\end{equation}

The transition function is unknown. Thus, we define the outputs of the transition function as samples from GPs, i.e.,
\begin{equation}
\tau_e(\myvec{s}, \myvec{a}) \sim \mathcal{GP}(0, k^{SE}_{\theta^{SE}_e}),
\label{eq:gp_transition}
\end{equation}
for each output feature $e$.

The expected long-term reward when following a policy $\pi$ is defined by a value function $V^\pi$. This function can be written recursively as the sum of the expected immediate reward and future reward, i.e.,
\begin{equation}
V^\pi(\mathbf{s}) = \mathbb{E}\left[R(\mathbf{s}, \pi(\myvec{s}), \mathbf{s'}) + \gamma V^\pi(\mathbf{s'})\ \middle|\ \myvec{s'} = \tau(\myvec{s}, \pi(\myvec{s}))\right].
\label{eq:vf}
\end{equation}
This is the sum of all possible long-term rewards weighted by their probability of occurrence when executing a policy $\pi$. The goal of an agent is then to find the optimal policy $\pi^*: S \to A$ which maximizes this expression. 

The expectation in Equation~\ref{eq:vf} typically has no closed-form expression for arbitrary reward and transition functions.
In order to approximate the value function, we use the Gaussian Process Reinforcement Learning (GPRL) method \cite{gprl}, which is a policy iteration method and thus iteratively evaluates and improves the policy $\pi$ until convergence. 

During the policy evaluation step, GPRL computes the values of a finite, but dense, vector of support points $\myvec{s}_\text{supp} = \langle\myvec{s}^{(i)}\rangle^N_{i=1}$. We use Latin hypercube sampling \cite{lhs_1979} to generate this vector, such that the state space is sufficiently covered. The values of these points can be computed analytically when the transition model and value function are described by a GP, and the reward function is bell-shaped.
Formally, given a policy $\pi$, a reward function centered around $\myvec{s}_\text{goal}$ with width $\sigma^2_R$, and an initial GP over the value function, the support values $\myvec{v}_\text{supp}$ have the recursive form:
\begin{equation}
\begin{split}
\myvec{v}_\text{supp} &= \myvec{r} + \gamma P \myvec{v}_\text{supp}\\
\mathrm{r}_i &= \frac{1}{\sqrt{|2\pi C^{(i)}|}}\exp\left(-\frac{1}{2} (\myvec{s}_\text{goal} - \boldsymbol{\mu}^{(i)})^T {C^{(i)}}^{-1} (\myvec{s}_\text{goal} - \boldsymbol{\mu}^{(i)})\right)\\
C^{(i)} &= \mathrm{\Sigma}^{(i)} + \sigma^2_R I,
\label{eq:gprl_evaluation}
\end{split}
\end{equation}
with the statistics of the transition model,
\begin{equation}
\begin{split}
\boldsymbol{\mu}^{(i)} &= \mathbb{E}\left[\myvec{s'}\ \middle|\ \myvec{s'} = \tau\left(\myvec{s}^{(i)}, \pi\left(\myvec{s}^{(i)}\right)\right)\right]\\
\mathrm{\Sigma}^{(i)} &= \V{\myvec{s'}\ \middle|\ \myvec{s'} = \tau\left(\myvec{s}^{(i)}, \pi\left(\myvec{s}^{(i)}\right)\right)},
\end{split}
\end{equation}
and $P$ a matrix that depends on the transition model and the value function.
The equation for the support values can be rewritten as a closed-form expression:
\begin{equation}
\begin{split}
\myvec{v}_\text{supp} &= (I - \gamma P)^{-1}\myvec{r}.
\end{split}
\end{equation}
We refer the reader to the work of Rasmussen and Kuss \cite{gprl} for more information about the exact form of the matrix $P$.

During the policy improvement step, a new GP is fitted over the value function $V(\cdot)$ using the support values to generalize over the state space. This function is used to optimize $\pi$:
\begin{equation}
\pi(\myvec{s}) \leftarrow \text{argmax}_{\myvec{a}} \mathbb{E}\left[R(\mathbf{s}, \myvec{a}, \mathbf{s'}) + \gamma V^\pi(\mathbf{s'})\ \middle|\ \myvec{s'} = \tau(\myvec{s}, \myvec{a})\right].
\label{eq:policy_improv}
\end{equation}
An expression similar to the one presented in Equation~\ref{eq:gprl_evaluation} can be obtained for arbitrary actions using the vector of support states. This expression can be maximized using standard stochastic optimizers for continuous action spaces or enumeration for discrete action spaces.

Throughout this manuscript, we only deal with deterministic transition models, and we thus set the observational noise $\sigma^2$ of the GP to $10^{-8}$ to ensure numerical stability. Other hyperparameters are optimized using evidence maximization on the training set \cite{intro_gps}.

\section{COREGIONALIZATION OVER MULTIPLE TRANSITION MODELS}
\label{sec:methods}
To transfer knowledge between fleet members, we leverage the statistical properties of GPs. Specifically, we assert that fleet members should only share information when they are correlated. This means that, for a given state-action, there exists a linear transformation between the outputs of the transition models of the two members.
Intuitively, the GP's covariance kernel allows to generalize in the regression model by correlating unobserved outputs to observed ones. Coregionalization extends this concept to the outputs of different GPs, suggesting that information from one process can be generalized to another. The main contribution in this work is the introduction of coregionalization to capture similarities between multiple transition models in order to decide whether and how knowledge should be transferred.

Formally, we define a fleet MDP as:
\begin{equation}
\begin{split}
\mathcal{M}_F &= (S, A, T^F, \gamma, R), \text{with}\\
&T^F = \{\tau_m\}^M_{m=1}.
\end{split}
\end{equation}
Compared to the definition in Section~\ref{sec:rl}, all properties are the same, except that $T^F$ is now a set of $M$ transition models, one for each fleet member $m$.

Consider a single member from the fleet, which we refer to as the \emph{target} and the rest of the fleet as the \emph{sources}. We denote the target's index as $t$, and the index of a source as $s$. In order to achieve knowledge transfer from the sources to the target, we must define a medium through which sources can communicate information. Specifically, for inputs $\myvec{x} = [\myvec{s}, \myvec{a}]$, we consider this model for the transition functions:
\begin{equation}
\begin{split}
&\tau_t(\myvec{x}) = \sum_{s \neq t} w_{t,s} g_s(\myvec{x}) + \alpha_t l_t(\myvec{x})\\
&\forall s \neq t:\\
&\qquad \tau_s(\myvec{x}) = w_{s,s} g_s(\myvec{x}) + \alpha_s l_s(\myvec{x}).\\
\end{split}
\label{eq:latent_function_model}
\end{equation}
The transition function of the target is a linear combination of $M-1$ global functions $g_s$ shared with every source $s$, and member-specific local functions (i.e., $l_t$ for the target and $l_s$ for the sources) to model the member's specific behavior. This ensures that all sources can exchange information with the target without compromising the specifics of a member's dynamics. The parameters $w_{t, s}$, $w_{s, s}$, $\alpha_{t}$ and $\alpha_{s}$ weigh the contribution of the different components in the transition functions. For example, when source $s$ has relevant information for the target $t$, the parameter $w_{t,s}$ should be high in order to transfer knowledge through function $g_s$. In contrast, when the sources have no relevant information for the target, the parameter $w_{t,s}$ should be zero in order to create independence between the source and the target.

We define the unknown components $g_s$, $l_s$ and $l_t$ as independent samples from a zero-mean GP with covariance kernel $k^{SE}_{\theta_{SE}}$ (see Equation~\ref{eq:gp_transition}). This entails that each of the transition functions is a linear combination of GP-distributed random variables, and is thus also a GP-distributed random variable \cite{intro_gps}. The mean functions of the transition models will be zero, due to the linearity of expectation property and the fact that all components have a mean of zero. Moreover, as the components are independently sampled, covariance can only exist within a single component, which is defined through the kernel $k^{SE}_{\theta_{SE}}$. The resulting cross-covariance equations for the transition functions are as follows:
\begin{equation}
\begin{split}
&\Cov{\tau_t(\myvec{x}), \tau_t(\myvec{x}')} = \left(\sum_{s\neq t} w^2_{t, s} + \alpha^2_{t}\right) k^{SE}_{\theta_{SE}}(\myvec{x}, \myvec{x}')\\
&\Cov{\tau_s(\myvec{x}), \tau_s(\myvec{x}')} = \left(w^2_{s,s} + \alpha^2_{s}\right) k^{SE}_{\theta_{SE}}(\myvec{x}, \myvec{x}')\\
&\Cov{\tau_t(\myvec{x}), \tau_s(\myvec{x}')} = \left(w_{t,s} w_{s,s}\right) k^{SE}_{\theta_{SE}}(\myvec{x}, \myvec{x}')\\
&\Cov{\tau_s(\myvec{x}), \tau_{s'}(\myvec{x}')} = 0,
\end{split}
\label{eq:cross_covs}
\end{equation}
where $s, s' \neq t$ and $s \neq s'$.

From these statistics, we can reformulate the target's transition function as a sample from a GP with the following fleet-wide kernel:
\begin{equation}
\begin{split}
k^F_{\theta^{(t)}}\left([\myvec{x}, m], [\myvec{x'}, m']\right) &= k^{SE}_{\theta_{SE}}(\myvec{x}, \myvec{x'})G_{m, m'}\\
G &= W + \left(\boldsymbol{\alpha}^2\right)^T I\\
W &= \sum_{s \neq t}\myvec{w}_s \myvec{w}^T_s
\label{eq:fleet_kernel}
\end{split}
\end{equation}
where $\myvec{w}_s$ only has non-zero elements at indices $t$ and $s$, and $m, m'$ are the indices of two fleet members. The matrix $W$ encodes relationships between the target and the sources, while $\boldsymbol{\alpha}$ contains independent terms for each member.

The decomposition of $G$ yields a valid covariance matrix, as it is symmetric positive semidefinite, i.e.,
\begin{equation}
\forall \myvec{z} \neq \myvec{0}\ :\ \myvec{z}^TG\myvec{z} = \sum_{s \neq t}||\myvec{z}^T\myvec{w}_s||\strut^2_2 + ||\myvec{z}^T\boldsymbol{\alpha}||\strut^2_2 \ge 0.
\end{equation}
The vector $\myvec{\theta}^{(t)}$ contains now both the hyperparameters $\myvec{\theta}_{SE}$ of the SE kernel and of the matrix $G$, i.e., $\myvec{w}$ and $\boldsymbol{\alpha}$. These parameters can be optimized using the training set $(X^F_{\text{tr}}, \myvec{y}^F_{\text{tr}})$ of the entire fleet, annotated with the indices of its members.

Note that the matrix $G$ contains $3M-2$ parameters per target (i.e., $M$ in $\boldsymbol{\alpha}$ and $2(M-1)$ in $W$). Therefore, the computational complexity is linear per target, and the method can be executed in a distributed manner over the fleet. Moreover, because of the sparsity of the defined covariances (see Equation~\ref{eq:cross_covs}), it is possible to significantly reduce the computational complexity of the matrix inversion in Equation~\ref{eq:post_normal}. Specifically, the computational complexity of the inversion operation can be reduced from $O\left(\left(\sum^M_{m=1} N_m\right)^3\right)$ to $O\left(\sum^M_{m=1} N^3_m\right)$, where $N_m$ is the number of samples of member $m$. The derivation of this complexity can be found in the Supplementary Information.\footnote{Supplementary Information: \url{https://drive.google.com/open?id=1EI5OZeHLh4AgXRwWx4j7CaPpPOmUbq4K}}
Both properties render the method scalable to larger fleets.

Using the new fleet covariance kernel, we can describe a single GP jointly over the outputs of all fleet members (Equation~\ref{eq:normal}) and compute the target's posterior statistics (Equation~\ref{eq:post_normal}) for regression. Note that even though the new model uses the whole fleet data set, the target can predict using its own transition function by computing the posterior statistics using index $t$, i.e.,
\begin{equation}
\boldsymbol{\tau}_t \ |\ X^{(t)}, X^F_{\text{tr}}, \myvec{y}^F_{\text{tr}}.
\end{equation}

We can define such a model for each member in the fleet independently by setting that member as the target, and thus construct the set $T^F_\theta$ by defining the transition model of each member $m$ as a fleet-wide GP:
\begin{equation}
\tau_m(\myvec{x}) \sim \mathcal{GP}\left(0, k^F_{\theta^{(m)}}\left([\myvec{x}, m], [\myvec{x'}, m']\right)\right).
\end{equation}

GPRL can be used for policy iteration to learn the optimal value function and policy. A high-level description of the complete fleet-wide policy iteration method for a given target is provided in Algorithm \ref{alg:fwpi}.

\begin{algorithm}
 \KwIn{Reward function $R$, set of support points $\myvec{s}_\text{supp}$, fleet-wide dynamics training data $(X^F_\text{tr}, \myvec{y}^F_\text{tr})$, target index $t$}
 \KwOut{Learned policy $\pi(\myvec{s})$}\nonl
 \  \\
 \textbf{Initialize:}\\
 \quad\ \ $\pi(\myvec{s}) \leftarrow $ random policy\;
 \quad\ \ $\myvec{v}_\text{supp} \leftarrow $ Apply reward function $R$ on $\myvec{s}_\text{supp}$\;
 \quad\ \ Define $V \sim \mathcal{GP}\left(0, k^{SE}_{\theta_V}\right)$;\\
 \quad\ \ Fit $V$ using $(S_\text{supp}, \myvec{v}_\text{supp})$;\\
 \textbf{Train fleet-wide transition model:} \hfill (Section 4)\\
 \quad\ \ $\theta^{(t)} \leftarrow \argmax\limits_{\theta_{SE}, G} p(\myvec{y}^F_\text{tr}\ |\ X^F_\text{tr}, \theta_{SE}, G)$;\\
 \quad\ \ Define $\tau \sim \mathcal{GP}(0, k^F_{\theta^{(t)}})$;\\
 \quad\ \ Fit $\tau$ using $(X^F_\text{tr}, \myvec{y}^F_\text{tr})$;\\
 \textbf{Policy iteration (GPRL):} \hfill (Section 2)\\
 \Indp\While{$\myvec{v}_\text{\upshape{supp}}$\upshape{ not converged}}{
  $\myvec{v}_\text{supp} \leftarrow$ Policy evaluation using $S_\text{supp}$, $R$, $\tau$, $V$ and $\pi$;\\
 Fit $V$ using $(S_\text{supp}, \myvec{v}_\text{supp})$;\\
 $\pi \leftarrow$ Policy improvement using $S_\text{supp}$, $R$, $\tau$ and $V$;\\
 }
 \caption{Fleet-Wide Policy Iteration}
 \label{alg:fwpi}
\end{algorithm}

\section{EXPERIMENTS}
\label{sec:experiments}

\begin{figure}[!t]
\begin{subfigure}{0.5\textwidth}
\centering
\includegraphics[scale=0.45,trim=0cm 0cm 0cm 0cm]{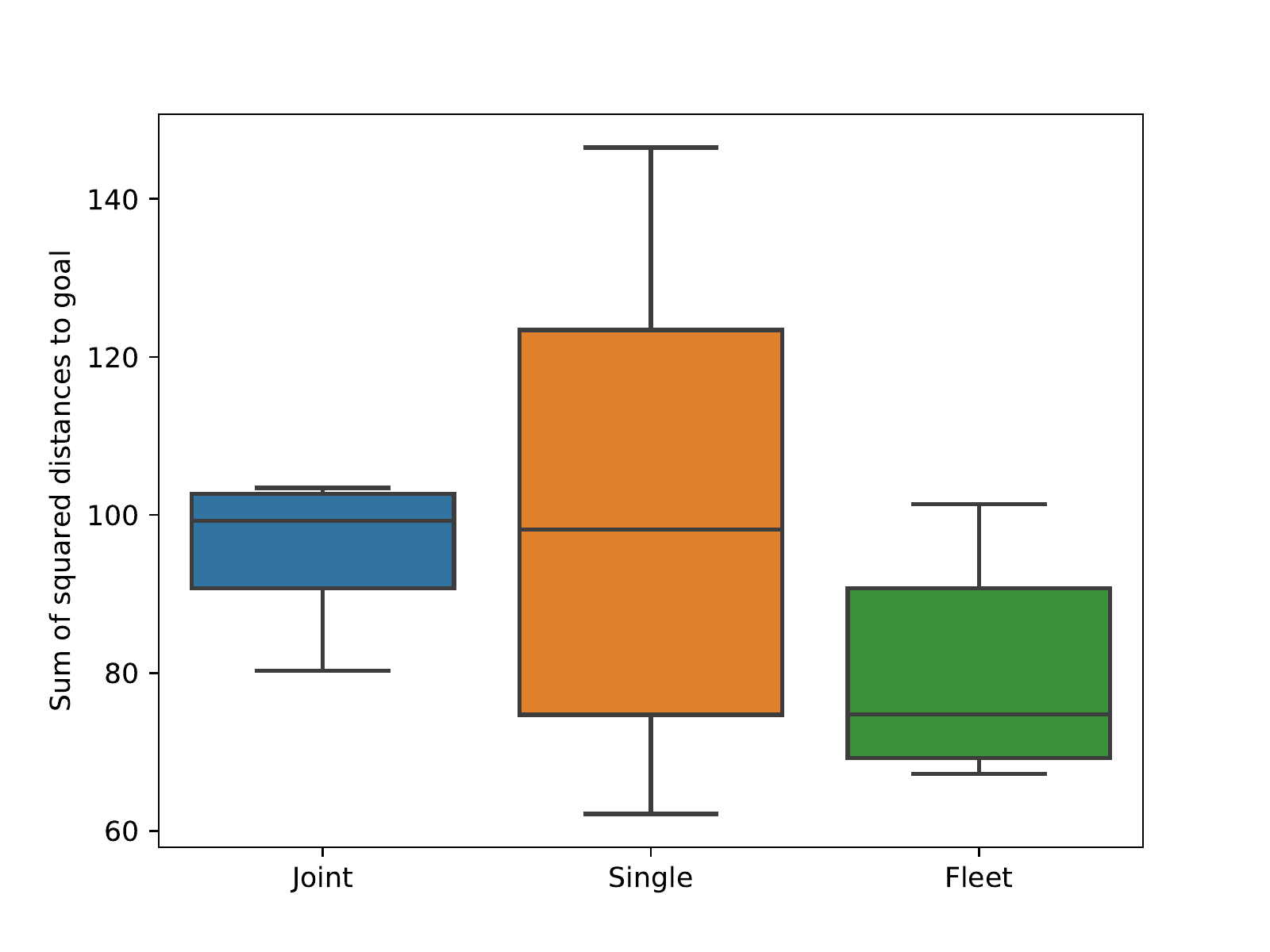}
\caption{Mountain car}
\label{fig:mc_si}
\end{subfigure}
\begin{subfigure}{0.5\textwidth}
\centering
\includegraphics[scale=0.45,trim=0cm 0cm 0cm 0cm]{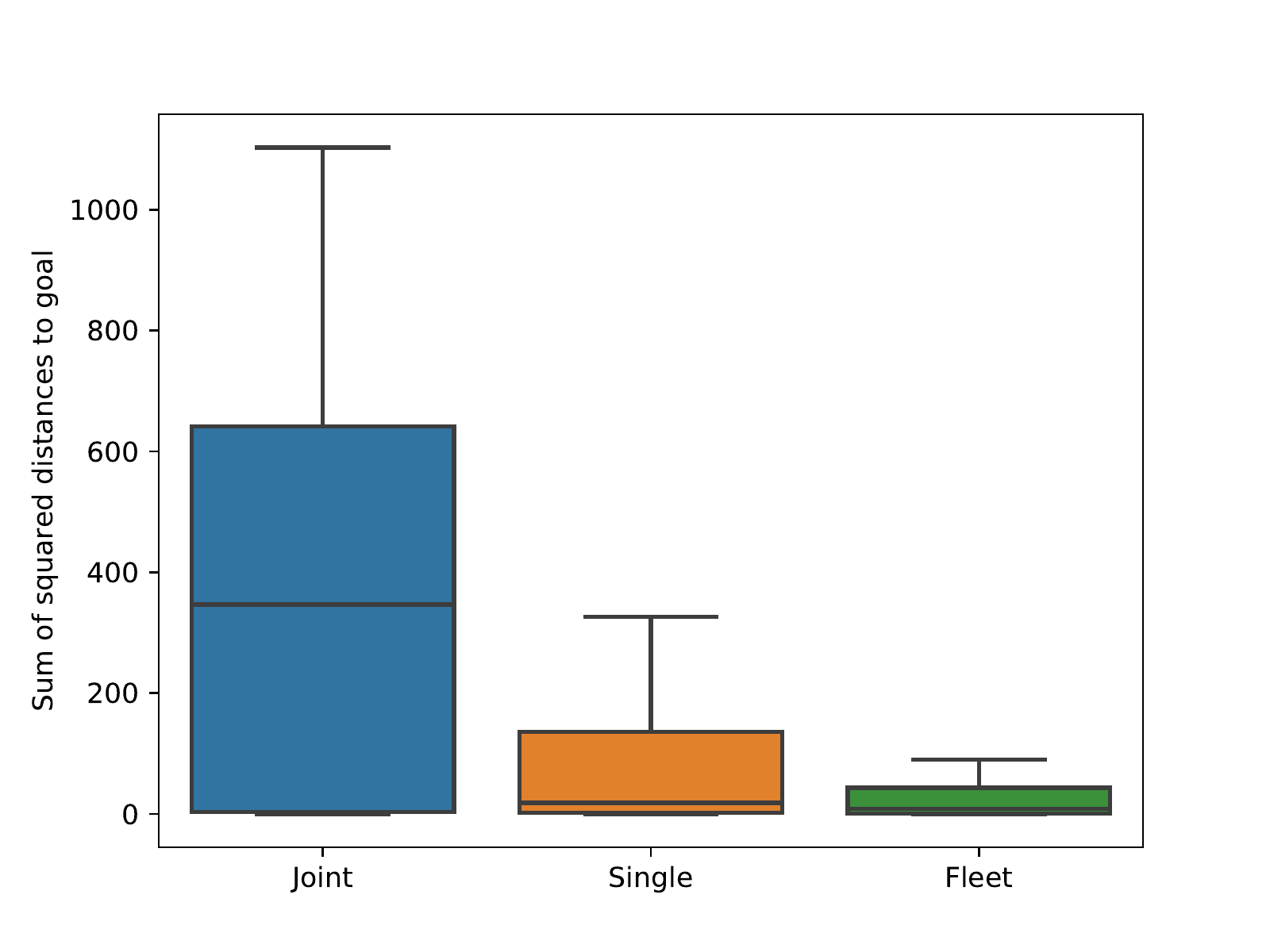}
\caption{Cart-Pole}
\label{fig:cp_si}
\end{subfigure}
\caption{Boxplot of the total sum of squared distances to the goal state for the mountain car (a) and cart-pole (b) benchmarks during 200 time steps. The experiment is repeated 50 times for each benchmark.}
\label{fig:mc_cp_results}
\end{figure}

First, we experimentally analyze our method on the well-known mountain car \cite{mountain_car_1994} and cart-pole \cite{barto1983neuronlike} benchmark problems. 
Next, we apply our method on a state-of-the-art wind farm simulator \cite{yaw_based_control_2017} to demonstrate our method's real-world benefits on larger fleets.\footnote{The source code of the method and experiments is available at \url{https://github.com/timo-verstraeten/fwpi-experiments}.}

In the first two benchmarks, we consider a fleet of 3 members. The fleet consists of a target, a similar source member A and a significantly different source member B. We sample the environments of sources A and B sufficiently, such that the dynamics are well-represented by the transition model of the respective fleet member. However, we provide the target with only a limited amount of data sampled from its own environment. This means that the target cannot sufficiently estimate its transition model based on these samples, making it challenging to find the optimal policy.
Therefore, transferring knowledge from member A will assist the target in finding the optimal policy. However, the dynamics in which member B operates are different from the target's dynamics, and sharing samples with it would misinform the target's transition model. Therefore, the objective of the target is to estimate a sufficiently accurate transition model by estimating the correlations with all sources and use the sources' knowledge proportional to the estimated correlation.
In the wind farm control task, we consider a fleet of 8 members. Again, we have a single target and sample the environments of the other fleet members.

Once the transition model is learned, we compute\footnote{We use the GPy library for all GP functionality \cite{gpy_2014}.} the optimal value function and policy using the GPRL method presented in Section~\ref{sec:rl}.
We consider an off-line batch RL setting and provide the learner with a random batch of transition samples.
The discount factor $\gamma$ is set to $0.99$ and the observational noise of the value GP is set to $0.1$.

In all experiments, we compare our method against two baselines, i.e., learning with a single target type and learning with a joint target type. The single target only uses its own samples to learn a transition model, while the joint target considers all fleet data jointly, assuming it is fully correlated with the sources. Specifically, we construct transition models that use the SE kernel described in Equation~\ref{eq:se_kernel}, fitted only on the target's own samples for the single target type, or using all fleet samples for the joint target type. For the fleet target type, we use our method to fit a transition model, using the fleet kernel described in Equation~\ref{eq:fleet_kernel}, based on all fleet samples.

\subsection{Mountain Car}

To illustrate our method, we set up the continuous mountain car domain \cite{mountain_car}. The car is positioned in a valley and its objective is to reach the top of the right-most hill. However, the slope is too steep for the car to simply accelerate to the top. Thus, it has to first drive up the opposite side of the valley and then accelerate from there to reach the top.

In this problem, a state $\myvec{s}$ consists of the position of the car (in $[-1.1, 0.55]$) and the velocity of the car (in $[-1, 1]$), while an action is a force applied to either side of the car (in $[-1, 1]$ times a power parameter). Both the state features and action are in the range $[-1, 1]$. The start and goal state are, respectively, given by $\myvec{s}_\text{start} = [-0.5, 0]$ and $\myvec{s}_\text{goal} = [0.45, 0]$, i.e., the bottom and top of the hill.
The standard deviation of the reward function $\sigma_R$ is set to 0.05. We use 200 support points.

We consider a fleet of three mountain cars: a target with a power of $1.5 \cdot 10^{-3}$ units, source A with power $10^{-3}$ and source B with power $10^{-4}$. For each source, we provide a batch of 100 transitions sampled uniformly random from its environment. We do the same for the target, but only sample 20 times, resulting in a total of 220 samples. We run the experiment 50 times for the three target types: single, joint and fleet.

We measure the performance of the methods by reporting the total sum of squared distances to the goal state during 200 time steps. The results are shown in Figure~\ref{fig:mc_si}. We observe that the joint target (i.e., learning from the full data set without the fleet kernel) rarely reaches the goal. This is because the target uses the data of source B, which has a low power parameter and is incapable of reaching the goal. Therefore, target does not expect to reach the goal and is often remaining at the bottom of the hill during runs. The single target (i.e., learning from own experiences) can sometimes achieve good results, but is unable to accurately represent its dynamics, due to the limited amount of data it can learn from. Because of the uncertainty in the transition model, the car is often incapable of finding a suitable policy. The fleet target consistently achieves good results, as the target is able to figure out which source is most useful to share data with through the fleet kernel.

\begin{figure*}[!h]
  \centering
  \mbox{
  \begin{subfigure}{0.33\textwidth}
    \includegraphics[scale=0.35]{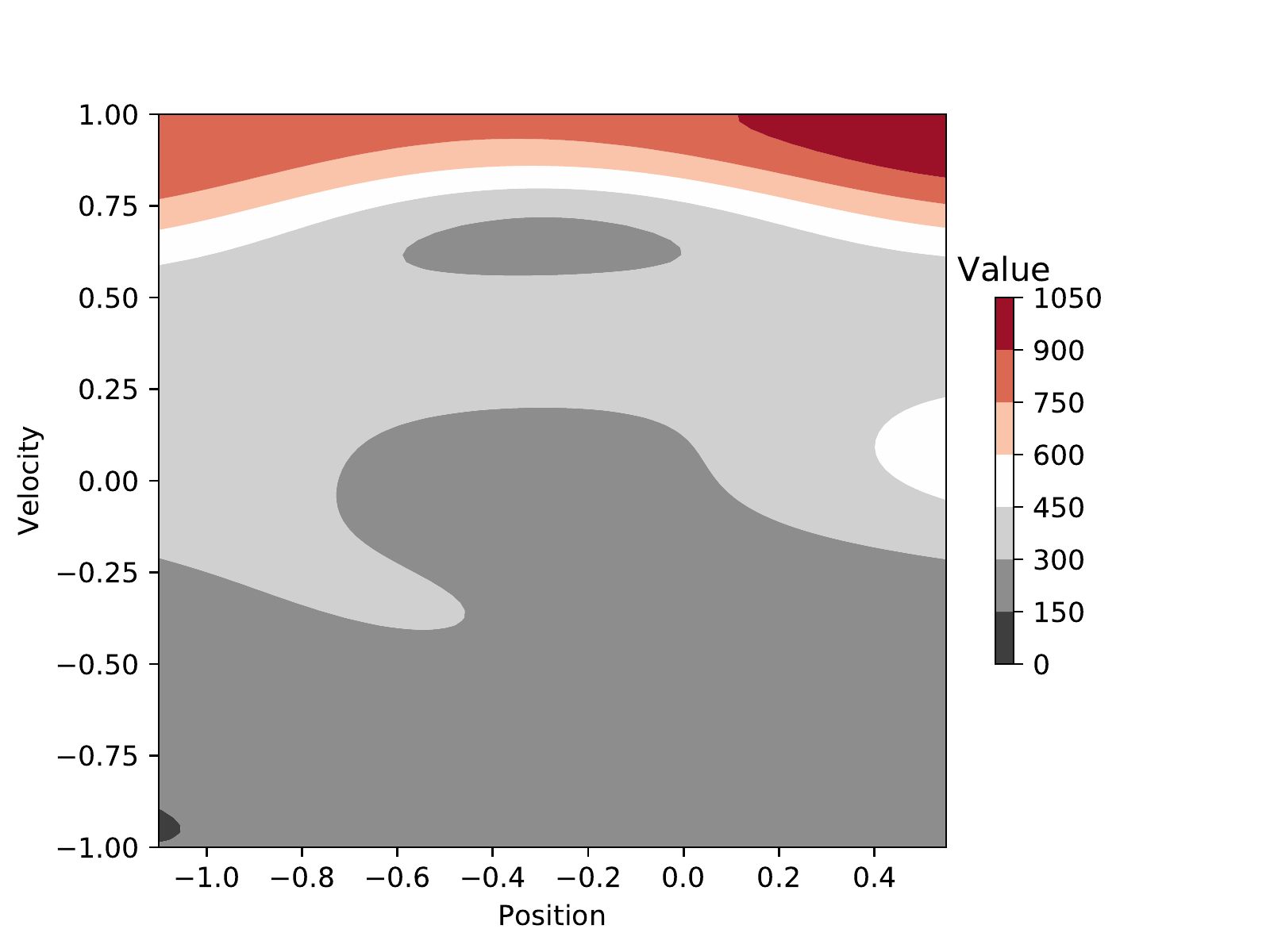}
    \caption{Single}
  \end{subfigure}
  \begin{subfigure}{0.33\textwidth}
    \includegraphics[scale=0.35,trim=0cm 0cm 0cm 0cm,clip]{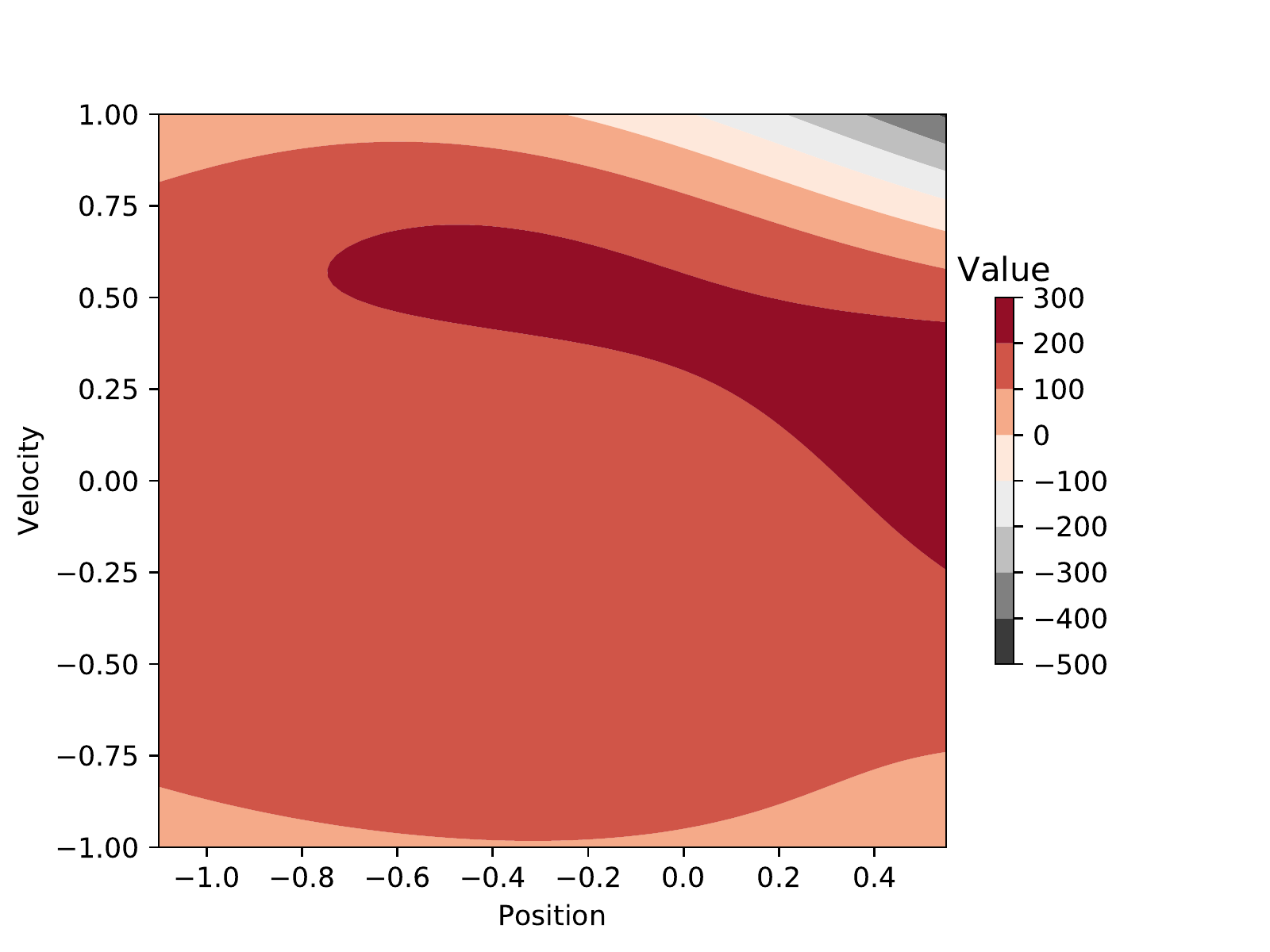}
    \caption{Joint}
  \end{subfigure}
  \begin{subfigure}{0.33\textwidth}
    \includegraphics[scale=0.35]{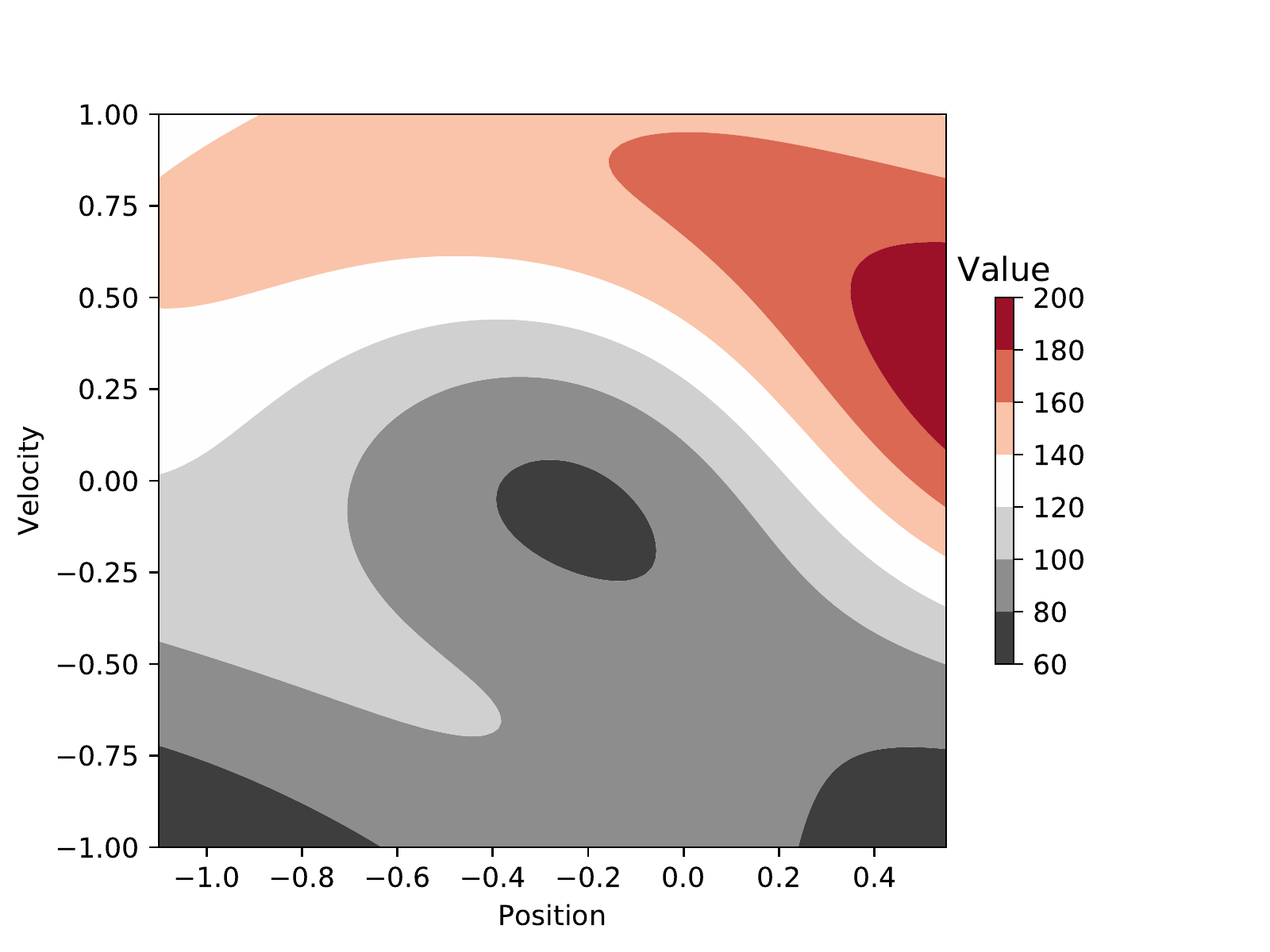}
    \caption{Fleet}
  \end{subfigure}
  }
  \caption{Mountain car -- Contour plots of the learned value functions (GPs) during the best runs of each target type. Each of the learner types are depicted; single (a), joint (b) and fleet (c) learning.}
  \label{fig:best_values}
\end{figure*}

In Figure~\ref{fig:best_values}, we plot the resulting GP of the value function during the best performant run for each of the target types. We can see that that the region with highest value matches the goal state for the fleet target, while the single and joint targets misidentify this region. The average standard deviation over the surface of the value GPs is 115 for the fleet target, 590 for the joint target and 3906 for the single target. This indicates that insufficient data is available to the single target to accurately represent the value function. In contrast, the fleet target has a low standard deviation, which means that it is certain about its value function.

\begin{figure}[!h]
\centering
\begin{subfigure}{0.49\textwidth}\centering
\includegraphics[scale=0.4, trim=1.1cm 0cm 3.7cm 0cm, clip]{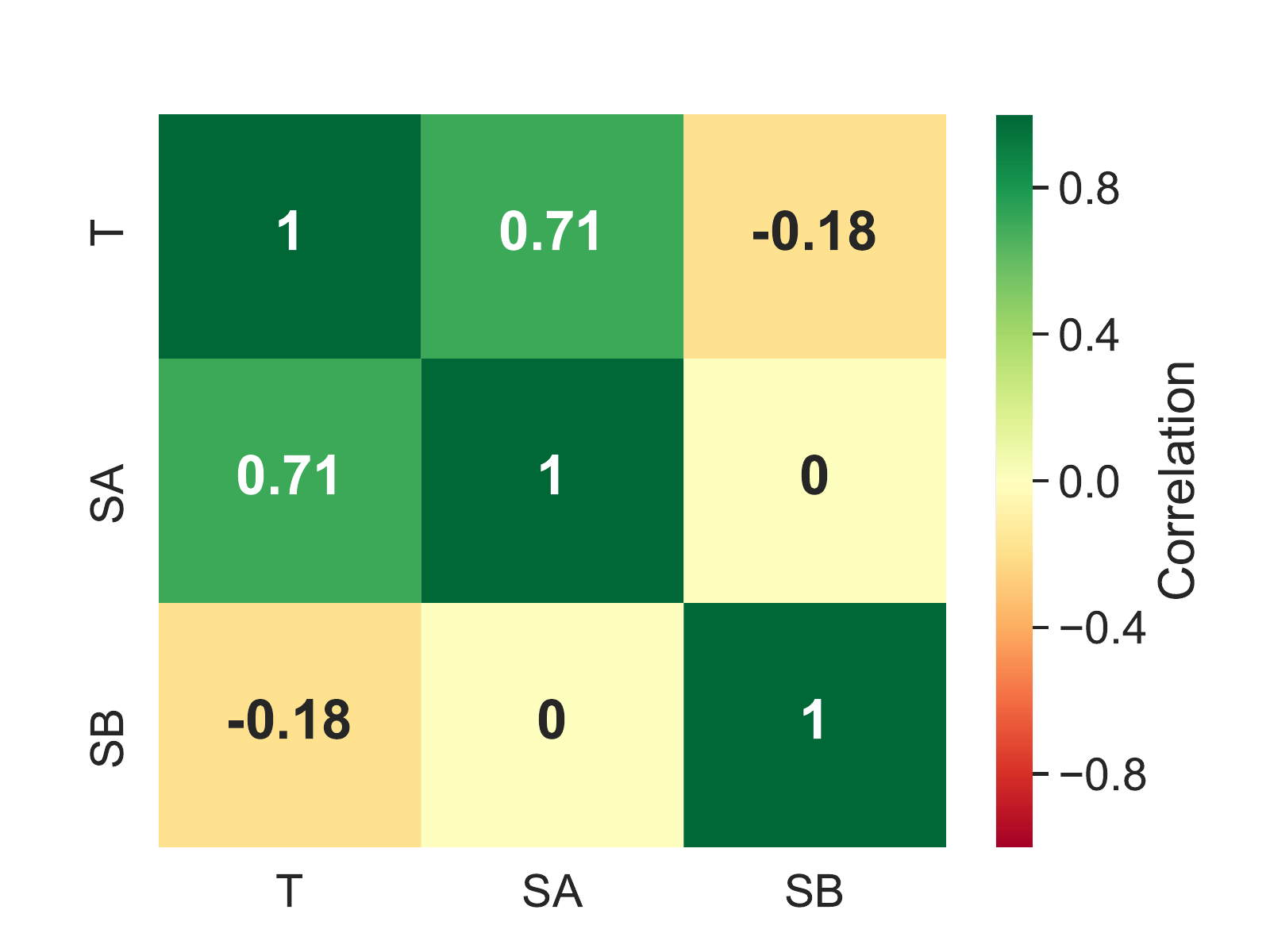}
\caption{Position}
\end{subfigure}%
\begin{subfigure}{0.49\textwidth}\centering
\includegraphics[scale=0.4, trim=1cm 0cm 0cm 0cm]{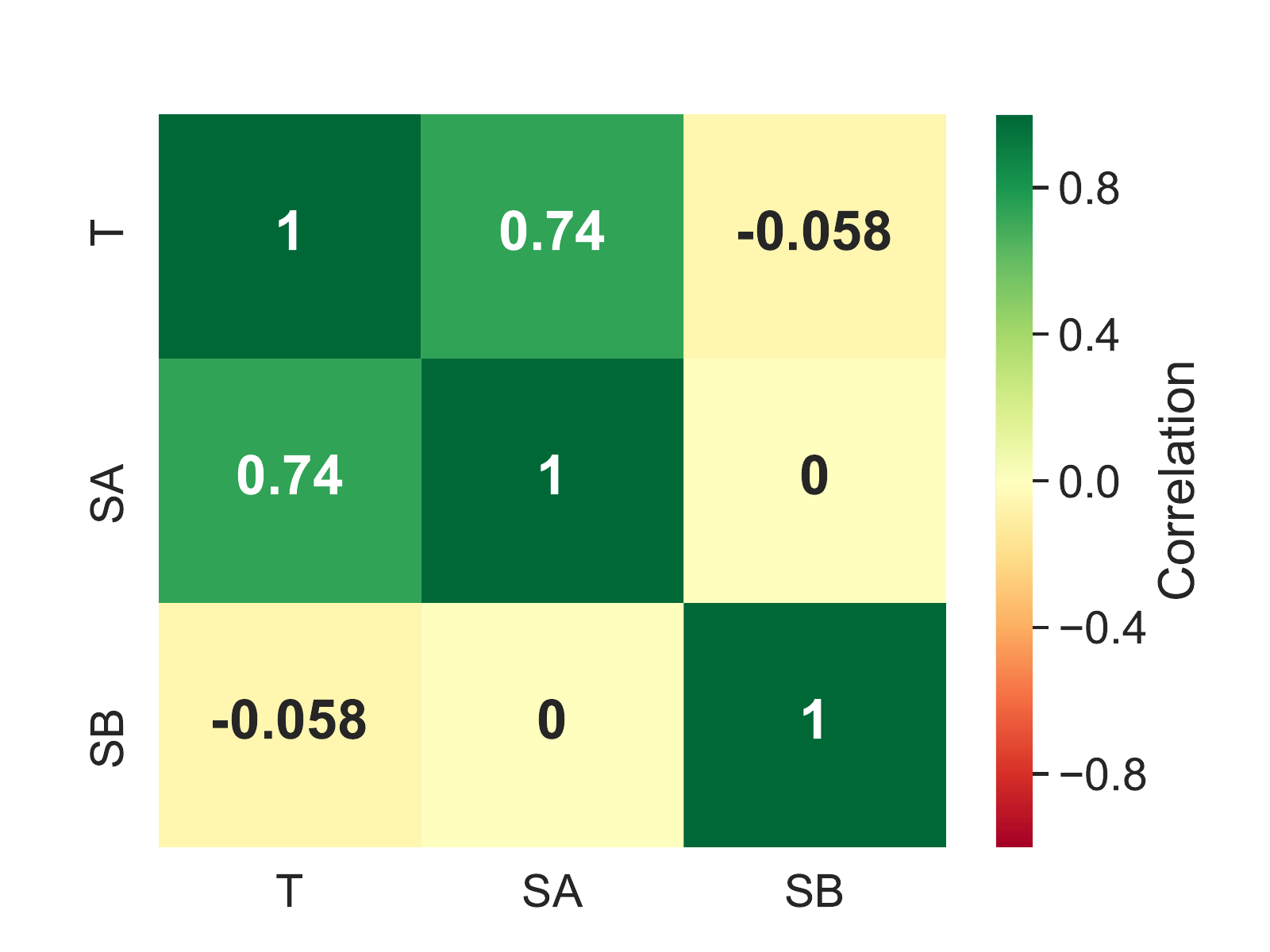}
\caption{Velocity}
\end{subfigure}
\caption{Mountain car -- Optimized correlation matrix between the fleet members, i.e., target (T), source A (SA) and source B (SB) for both state dimensions.}
\label{tab:corr}
\end{figure}

Next, we plot the correlation matrices learned by the fleet target, averaged over all runs. Given the optimized cross-covariance matrix $G$ from Equation~\ref{eq:fleet_kernel}, we can compute the correlation matrix:
\begin{equation}
\text{corr}(G) = (\text{diag}(G))^{-0.5} G (\text{diag}(G))^{-0.5}.
\end{equation}
The element-wise means of these matrices over all runs are given in Figure~\ref{tab:corr}. The fleet target successfully identifies source A to be similar, while assigning a notably lower correlation value to source B.

\subsection{Cart-Pole}

In the cart-pole domain, the goal is to keep a pole balanced on top of a controllable cart. Cart-pole is an underactuated system, as the system has more degrees of freedom than actuators. Balancing the pole is challenging, as its equilibrium is highly unstable.

In this problem, a state $\myvec{s}$ consists of the position of the cart (in $[-4.8, 4.8]$), the angular position of the pole (in $[-0.42, 0.42]$), the velocity of the cart (in $[-2, 2]$) and the angular velocity of the pole (in $[-2, 2]$). The start and goal state are the same, namely, at the equilibrium, i.e., $\myvec{s}_\text{start} =\myvec{s}_\text{goal} = [0, 0]$.
The standard deviation of the reward function $\sigma_R$ is set to 0.2. We set the number of support points to 300.

We consider a fleet of three carts: a target with a pole mass of $0.1$ units, source A with mass $0.2$ and source B with mass $0.5$. For each source, we provide a batch of 50 transitions sampled uniformly random from its environment. We do the same for the target, but only sample 5 times, resulting in a total of 105 samples. We run the experiment 50 times for the three target types: single, joint and fleet.

We measure the performance of the methods by reporting the total sum of squared distances from the equilibrium during 200 time steps. The results are shown in Figure~\ref{fig:cp_si}. Due to the instability of the equilibrium, it is necessary to accurately represent the transition model. The joint target fails to achieve this, as it aggregates samples over different dynamics. The single target achieves better results, but is often uncertain about its transition model, leading to suboptimal behavior. The fleet target consistently manages to keep the pole around its equilibrium.

\subsection{Wind Turbines}
\label{sec:experiments_wind}

To demonstrate the need for our method in practical applications, we introduce a new setting in the context of wind energy. Current wind turbine controllers position the rotors toward the measured incoming wind vector to ensure high productivity \cite{boersma2017tutorial}. However, as wind passes through upstream turbines, the wind speed is reduced and the energy extracted by downstream turbines is significantly lower, which is referred to as the wake effect. When considering wind farms (i.e., groups of wind turbines), it is essential to take this effect in consideration \cite{wake_2012}.

In recent work, steering wake through rotor misalignment is investigated \cite{yaw_based_control_2017}.
For example, in a setting with two wind turbines, the upstream turbine slightly misaligns its rotor to deflect the wake away from the downstream turbine. Therefore, although the upstream turbine itself produces less energy, the group's total productivity is increased.

Because of the complexity of the wake effect and incomplete knowledge about a turbine's condition, it is necessary to gather data in the field about potential control policies, rendering it a reinforcement learning problem. As learning policies from scratch could result in potential revenue loss, fleet-wide policy iteration can improve the learning speed. Moreover, our batch RL setting makes sense, as wind farm service providers first need to thoroughly assess the performance of the acquired policy before implementing it \cite{boersma2017tutorial}.

We demonstrate our method on a fleet of two-turbine rows in a wind farm that consists of 8 rows and show how information exchange between transition models can improve the learning speed. We use the state-of-the-art open source WISDEM FLORIS simulator to model the wind farm dynamics and wake effect \cite{FLORIS_2019}, and use the 5-megawatt (MW) reference turbine description from the National Renewable Energy Laboratory to model the individual wind turbines \cite{nrel_turbine_def_2009}. 

In this environment, the state consists of the orientations of both turbines (values in $[-45, 45]$ degrees with respect to the wind vector) and the associated total power production (values in $[0.5, 1.05]$ MW). The actions are changes in orientation with values of either -1, 0 or 1 degrees. The start state is both turbines aligned with the wind vector, which is current practice in wind turbine control \cite{boersma2017tutorial}. The goal $\myvec{s}_\text{goal}$ is centered around a power production of $1.07$ MW with a scale $\sigma_R$ of $0.05$ to encourage high productivity. The number of support points is set to 300.

Each two-turbine row represents a fleet member. Again, we report the results for one target. This is considered to be a new row of which the generator efficiency is set to 1. However, we set the generator efficiencies to 0.9 for 3 source members, and to 0.8 for the remaining 4 source members, which is a realistic configuration that could be the result of aging \cite{degradation_wind_farm_2014}. The turbines are positioned 100 m apart on a line parallel to the incoming wind vector. The wind speed is set to 6 m/s. We assume independence between turbine rows, which is a reasonable assumption given the specified wind vector, since wake generated by one row will not influence the other turbine rows.

\begin{figure}[!h]
\centering
\includegraphics[scale=0.5,trim=0cm 0cm 0cm 0cm]{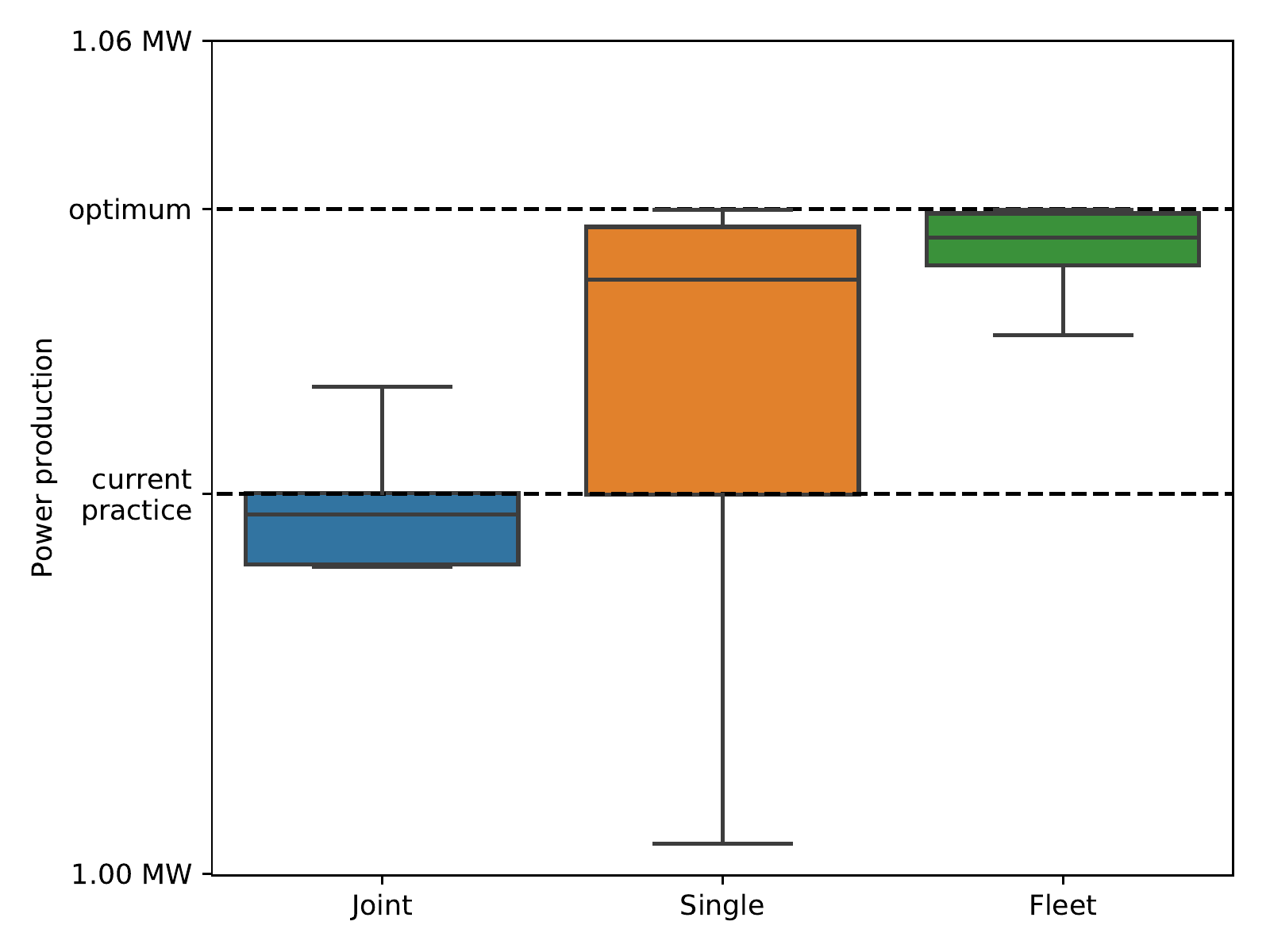}
\caption{Wind Farm -- Boxplot of the power productions for each target type over 50 runs. The optimal performance, as well as the performance achieved when using the control policy used in current practice, are given (dashed lines).}
\label{fig:wind_perf}
\end{figure}

To each turbine row, we provide a batch of 50 transitions randomly sampled from its environment. We measure the performance of the methods by reporting the power production (MW) achieved at the end of the run. We compare the targets to the performance achieved under the optimal policy and to the performance under the policy used in current practice, i.e., aligning all turbines with the incoming wind vector. The results are shown in Figure~\ref{fig:wind_perf}.

To each member, we provide a batch of 50 transitions sampled uniformly random from their environment, resulting in 400 fleet samples. We run the experiment 50 times for each of the target types: single, joint and fleet. We show the results in Figure~\ref{fig:wind_perf}. 

We observe that the single target has a wide variance on its performance. The uncertainty about its transition model is high, due to the limited amount of data it has access to. The joint target has lower variance, but has the worst performance, close to the performance of current practice policies. As all data is aggregated, many transition samples are not representative for the true dynamics of the target. The fleet target consistently achieves results that closely follow the optimal performance, as it has the ability to differentiate between relevant and irrelevant samples over the entire fleet data set.


\section{DISCUSSION}

Fleet-wide policy iteration outperforms the two baselines over all of our experiments.
The joint target often fails to reach the goal, while the single target is often uncertain about its own transition model. This reflects a trade-off in bias and variance, where we have to decide to either use all data at the risk of misrepresenting the transition model (bias), or only use representative data while remaining uncertain about the model (variance). Our method successfully balances both by properly weighing each source with their correlation with the target. This is reflected in the learned value functions. The fleet target finds the region of highest reward, which is around the goal state, while the single and joint targets misrepresent their value function.
We further validated the ability of our method to balance between bias and variance through a sensitivity analysis on the mountain car setting. Specifically, we varied the power parameter of source A between $0.0005$ and $0.0015 \cdot 10^{-4}$ to simulate a range of similarities between source A and the target. The fleet target outperforms both baselines, and exhibits similar performance to the single target when the target is significantly different from source A, and thus no information transfer is possible. More information on this analysis can be found in the Supplementary Information.

The successful use of data exchange in fleets has strong implications for the real world applications. The wind farm experiment shows that close-to-optimal performance can be achieved when using fleet-wide policy iteration, while the alternatives (i.e., single and joint learning) often yield performances close to current practice or worse.

The fleet-wide transition model is a sparse variant of the intrinsic coregionalization model (ICM) \cite{geo_coreg,multitask_kernel}. The ICM captures cross-covariances between multiple functions, and thus improves the accuracy of those functions jointly. However, as we consider multiple sources and a single target, the target's transition model will be tailored toward improving its own accuracy, rather than the joint accuracy over all fleet members. Additionally, the computational burden when using our sparse coregionalization model is significantly lower. Our method can be executed in a distributed manner over the fleet, which reduces the quadratic complexity of the coregionalization matrix in the ICM model, to a linear complexity per target member. Moreover, as the covariances in Equation~\ref{eq:cross_covs} are sparse, the inversion operation can be made linear in the number of fleet members as well. This renders our method scalable to larger fleets.

In future work, we will further improve the scalability of our method by using sparse GP approximations \cite{snelson2006sparse,wilson2015kernel} to reduce the computational burden of the matrix inversion. As many of these methods are independent with respect to the covariance kernel or require a factorable kernel, our model is extensible to many of these approximations. By using a sparse GP approximation, our fleet RL method can handle even larger data sets and fleets.


\section{CONCLUSION}
\label{sec:conclusion}

In this work, we introduced a novel sample-efficient fleet reinforcement learning method, called \emph{fleet-wide policy iteration}, based on Gaussian processes and coregionalization. It estimates cross-covariances between a fleet of machines and transfers knowledge between them.

We provided experimental results on two benchmark problems: mountain car and cart-pole. In these settings, a target has to share data with a similar and dissimilar fleet member. While the two baselines, i.e., no transfer and single model learning, perform poorly, the learner that uses our fleet-wide policy iteration method manages to solve both tasks consistently. Additionally, we provided a real-world example of how fleet-wide policy iteration can be used in wind farms. The method shows overall improvement in terms of the wind farm's productivity.

\subsection*{Acknowledgments}
The authors would like to acknowledge FWO (Fonds Wetenschappelijk Onderzoek) for their support through the SB grant of Timothy Verstraeten (1S47617N) and the SB grant of Pieter JK Libin (1S31916N). 

\bibliographystyle{abbrv}
\bibliography{refs}  

\begin{thebibliography}{10}

\bibitem{barto1983neuronlike}
A.~G. Barto, R.~S. Sutton, and C.~W. Anderson.
\newblock Neuronlike adaptive elements that can solve difficult learning
  control problems.
\newblock {\em IEEE transactions on systems, man, and cybernetics},
  (5):834--846, 1983.

\bibitem{boersma2017tutorial}
S.~Boersma, B.~Doekemeijer, P.~M. Gebraad, P.~A. Fleming, J.~Annoni, A.~K.
  Scholbrock, J.~Frederik, and J.-W. van Wingerden.
\newblock A tutorial on control-oriented modeling and control of wind farms.
\newblock In {\em 2017 American Control Conference (ACC)}, pages 1--18. IEEE,
  2017.

\bibitem{multitask_kernel}
E.~V. Bonilla, K.~M.~A. Chai, and C.~K.~I. Williams.
\newblock Multi-{Task} {Gaussian} {Process} {Prediction}.
\newblock {\em Advances in Neural Information Processing Systems}, 20:153--160,
  2008.

\bibitem{marl_survey_2006}
L.~{Busoniu}, R.~{Babuska}, and B.~{De Schutter}.
\newblock Multi-agent reinforcement learning: A survey.
\newblock In {\em The 9th International Conference on Control, Automation,
  Robotics and Vision}, pages 1--6, 2006.

\bibitem{deisenroth2014multi}
M.~P. Deisenroth, P.~Englert, J.~Peters, and D.~Fox.
\newblock Multi-task policy search for robotics.
\newblock In {\em IEEE International Conference on Robotics and Automation
  (ICRA)}, pages 3876--3881. IEEE, 2014.

\bibitem{pilco}
M.~P. Deisenroth and C.~E. Rasmussen.
\newblock {PILCO}: A model-based and data-efficient approach to policy search.
\newblock In {\em Proc. of the International Conference on Machine Learning
  (ICML)}. 2011.

\bibitem{hidmdp_2016}
F.~Doshi-Velez and G.~Konidaris.
\newblock Hidden parameter markov decision processes: A semiparametric
  regression approach for discovering latent task parametrizations.
\newblock In {\em Proc. of the 25th International Joint Conference on
  Artificial Intelligence (IJCAI)}, pages 1432--1440, 2016.

\bibitem{gptd}
Y.~Engel, S.~Mannor, and R.~Meir.
\newblock {Reinforcement Learning with Gaussian Processes}.
\newblock In {\em Proc. of the 22nd International Conference on Machine
  Learning}, pages 201--208. ACM Press, 2005.

\bibitem{yaw_based_control_2017}
P.~Gebraad, J.~J. Thomas, A.~Ning, P.~Fleming, and K.~Dykes.
\newblock Maximization of the annual energy production of wind power plants by
  optimization of layout and yaw-based wake control.
\newblock {\em Wind Energy}, 20(1):97--107, 2017.

\bibitem{fleet_vehicles_2014}
M.~Gerla, E.-K. Lee, G.~Pau, and U.~Lee.
\newblock Internet of vehicles: {F}rom intelligent grid to autonomous cars and
  vehicular clouds.
\newblock In {\em 2014 IEEE world forum on internet of things (WF-IoT)}, pages
  241--246. IEEE, 2014.

\bibitem{wake_2012}
F.~Gonz{\'a}lez-Longatt, P.~Wall, and V.~Terzija.
\newblock Wake effect in wind farm performance: Steady-state and dynamic
  behavior.
\newblock {\em Renewable Energy}, 39(1):329--338, 2012.

\bibitem{geo_coreg}
P.~Goovaerts.
\newblock Geostatistics for natural resource evaluation.
\newblock 42, 1997.

\bibitem{gpy_2014}
{GPy}.
\newblock {GPy}: A gaussian process framework in python.
\newblock \url{http://github.com/SheffieldML/GPy}, since 2012.

\bibitem{fleet_wide_condition_monitoring_2018}
J.~Helsen, C.~Peeters, T.~Verstraeten, J.~Verbeke, N.~Gioia, and A.~Now{\'e}.
\newblock Fleet-wide condition monitoring combining vibration signal processing
  and machine learning rolled out in a cloud-computing environment.
\newblock In {\em International Conference on Noise and Vibration Engineering
  (ISMA)}, 2018.

\bibitem{ijspeert2003learning}
A.~J. Ijspeert, J.~Nakanishi, and S.~Schaal.
\newblock Learning attractor landscapes for learning motor primitives.
\newblock In {\em Advances in neural information processing systems}, pages
  1547--1554, 2003.

\bibitem{nrel_turbine_def_2009}
J.~Jonkman, S.~Butterfield, W.~Musial, and G.~Scott.
\newblock Definition of a 5-{MW} reference wind turbine for offshore system
  development.
\newblock Technical report, National Renewable Energy Laboratory (NREL),
  Golden, CO, USA, 2009.

\bibitem{hidden_param_mdp_2017}
T.~W. Killian, S.~Daulton, G.~Konidaris, and F.~Doshi-Velez.
\newblock Robust and efficient transfer learning with hidden parameter markov
  decision processes.
\newblock In I.~Guyon, U.~V. Luxburg, S.~Bengio, H.~Wallach, R.~Fergus,
  S.~Vishwanathan, and R.~Garnett, editors, {\em Advances in Neural Information
  Processing Systems (NIPS)}, volume~30, pages 6250--6261. Curran Associates,
  Inc., 2017.

\bibitem{kober2011reinforcement}
J.~Kober, E.~Oztop, and J.~Peters.
\newblock Reinforcement learning to adjust robot movements to new situations.
\newblock In {\em Proc. of the 22nd International Joint Conference on
  Artificial Intelligence (IJCAI)}, 2011.

\bibitem{konidaris2012transfer}
G.~Konidaris, I.~Scheidwasser, and A.~Barto.
\newblock Transfer in reinforcement learning via shared features.
\newblock {\em Journal of Machine Learning Research (JMLR)}, 13:1333--1371,
  2012.

\bibitem{bayesian_mtrl_value}
A.~Lazaric and M.~Ghavamzadeh.
\newblock Bayesian multi-task reinforcement learning.
\newblock In {\em Proc. of the 27th International Conference on Machine
  Learning (ICML)}, pages 599--606. Omnipress, 2010.

\bibitem{wind_farms_fleet_2016}
R.~Martin, I.~Lazakis, S.~Barbouchi, and L.~Johanning.
\newblock Sensitivity analysis of offshore wind farm operation and maintenance
  cost and availability.
\newblock {\em Renewable Energy}, 85:1226--1236, 2016.

\bibitem{lhs_1979}
M.~D. McKay, R.~J. Beckman, and W.~J. Conover.
\newblock Comparison of three methods for selecting values of input variables
  in the analysis of output from a computer code.
\newblock {\em Technometrics}, 21(2):239--245, 1979.

\bibitem{mountain_car}
A.~Moore.
\newblock {\em Efficient Memory-Based Learning for Robot Control}.
\newblock PhD thesis, University of Cambridge, 1990.

\bibitem{mountain_car_1994}
A.~W. Moore.
\newblock The parti-game algorithm for variable resolution reinforcement
  learning in multidimensional state-spaces.
\newblock In {\em Proc. of the 6th International Conference on Neural
  Information Processing Systems (NIPS)}, pages 711--718, 1993.

\bibitem{FLORIS_2019}
NREL.
\newblock {FLORIS. Version 1.0.0}, 2019.

\bibitem{pan2009survey}
S.~J. Pan and Q.~Yang.
\newblock A survey on transfer learning.
\newblock {\em IEEE Transactions on knowledge and data engineering},
  22(10):1345--1359, 2009.

\bibitem{mdp_definition}
M.~L. Puterman.
\newblock {\em {Markov Decision Processes: Discrete Stochastic Dynamic
  Programming}}.
\newblock John Wiley \& Sons, Inc., 1st edition, 1994.

\bibitem{gprl}
C.~E. Rasmussen and M.~Kuss.
\newblock Gaussian processes in reinforcement learning.
\newblock 16:751--758, 2003.

\bibitem{intro_gps}
C.~E. Rasmussen and C.~K.~I. Williams.
\newblock {\em Gaussian {Processes} for {Machine} {Learning}}.
\newblock The {MIT} Press, Cambridge, MA, USA, 2006.

\bibitem{meta_rl_2018}
S.~S{\ae}mundsson, K.~Hofmann, and M.~P. Deisenroth.
\newblock Meta reinforcement learning with latent variable gaussian processes.
\newblock In {\em Proc. of the 34th Conference on Uncertainty in Artificial
  Intelligence (UAI)}, pages 642--652, 2018.

\bibitem{snelson2006sparse}
E.~Snelson and Z.~Ghahramani.
\newblock Sparse gaussian processes using pseudo-inputs.
\newblock In {\em Advances in neural information processing systems}, pages
  1257--1264, 2006.

\bibitem{degradation_wind_farm_2014}
I.~Staffell and R.~Green.
\newblock How does wind farm performance decline with age?
\newblock {\em Renewable energy}, 66:775--786, 2014.

\bibitem{intro_rl}
R.~S. Sutton and A.~G. Barto.
\newblock {\em Reinforcement learning: An introduction}.
\newblock MIT press Cambridge, 1998.

\bibitem{taylor2007cross}
M.~E. Taylor and P.~Stone.
\newblock Cross-domain transfer for reinforcement learning.
\newblock In {\em Proc. of the 24th International Conference on Machine
  Learning (ICML)}, pages 879--886. ACM, 2007.

\bibitem{verstraeten2019fleetwide}
T.~Verstraeten, A.~Nowe, J.~Keller, Y.~Guo, S.~Sheng, and J.~Helsen.
\newblock Fleetwide data-enabled reliability improvement of wind turbines.
\newblock {\em Renewable and Sustainable Energy Reviews}, 109:428--437, 2019.

\bibitem{bayesian_mtrl_mdp}
A.~Wilson, A.~Fern, S.~Ray, and P.~Tadepalli.
\newblock Multi-task reinforcement learning: A hierarchical bayesian approach.
\newblock In {\em Proc. of the 24th International Conference on Machine
  Learning (ICML)}, pages 1015--1022, 2007.

\bibitem{wilson2015kernel}
A.~Wilson and H.~Nickisch.
\newblock Kernel interpolation for scalable structured gaussian processes
  ({KISS-GP}).
\newblock In {\em International Conference on Machine Learning (ICML)}, pages
  1775--1784, 2015.

\bibitem{sample_efficiency_2018}
Y.~Yu.
\newblock Towards sample efficient reinforcement learning.
\newblock In {\em Proc. of the 27th International Joint Conference on
  Artificial Intelligence (IJCAI)}, pages 5739--5743. International Joint
  Conferences on Artificial Intelligence Organization, 2018.

\end{thebibliography}

\end{document}